\documentclass[letterpaper,10pt,conference]{ieeeconf}

\IEEEoverridecommandlockouts
\usepackage{amsmath}
\usepackage{amssymb}
\usepackage{booktabs}
\usepackage{multirow}
\usepackage{colortbl}
\usepackage{arydshln}
\usepackage{cite}
\usepackage{graphicx}
\usepackage{url}
\usepackage{placeins}
\makeatletter
\let\NAT@parse\@undefined
\makeatother
\usepackage[colorlinks,allcolors=black,citecolor=blue]{hyperref}

\makeatletter
\newcommand*{\memx@bracketedcite}[1]{%
  \begingroup
    \let\memx@originalcite@link\hyper@@link
    \def\hyper@@link[##1]##2##3##4{%
      \memx@originalcite@link[##1]{##2}{##3}{[##4]}%
    }%
    #1%
  \endgroup
}
\let\memx@original@citex\@citex
\def\@citex[#1]#2{%
  \if@tempswa
    \memx@original@citex[#1]{#2}%
  \else
    \begingroup
      \let\citeleft\@empty
      \let\citeright\@empty
      \renewcommand*{\citepunct}{, }%
      \renewcommand*{\citedash}{--}%
      \let\citeform\memx@bracketedcite
      \memx@original@citex[]{#2}%
    \endgroup
  \fi
}
\makeatother

\title{\LARGE \bf
Mem-X: A Memory-Dependent Cross-Episode Benchmark for
Vision-and-Language Navigation
}

\author{%
\authorblockN{%
Zhikang Shi\textsuperscript{1},
YUK TUNG SAMUEL FANG\textsuperscript{1},
Hongyu Ding\textsuperscript{2},
Zixuan Chen\textsuperscript{1},
Yujin Wang\textsuperscript{1},\\
Junhui Guo\textsuperscript{1},
You Wu\textsuperscript{1},
Jieqi Shi\textsuperscript{1,*},
Jing Huo\textsuperscript{2},
Yang Gao\textsuperscript{1}}
\authorblockA{%
\textsuperscript{1}School of Intelligence Science and Technology, Nanjing University, China.\\
\textsuperscript{2}School of Computer Science, Nanjing University, China.\\
\textsuperscript{*}Corresponding author.}
}

\hypersetup{%
  pdflinkmargin=0pt,
  pdftitle={Mem-X: A Memory-Dependent Cross-Episode Benchmark for Vision-and-Language Navigation},
  pdfauthor={Zhikang Shi, YUK TUNG SAMUEL FANG, Hongyu Ding, Zixuan Chen, Yujin Wang, Junhui Guo, You Wu, Jieqi Shi, Jing Huo, Yang Gao}
}

\begin{document}

\maketitle
\thispagestyle{empty}
\pagestyle{empty}

\begin{abstract}
Long-term memory enables embodied agents to reuse past observations across tasks. However, existing cross-episode vision-and-language navigation (VLN) benchmarks often leave it unclear whether successful navigation relies on historical memory or current exploration. We introduce Mem-X, a benchmark designed to evaluate memory-dependent cross-episode VLN. Mem-X organizes episodes into sessions in which subsequent instructions refer to objects or routes encountered in earlier episodes. Its three levels evaluate historical-object recall, in-memory localization and episodic indexing, and off-memory relocalization and historical-route reproduction. For the latter two levels, agents receive episode-local poses but no transformations between episode frames, requiring them to establish spatial correspondence with historical memory. Mem-X comprises 2,071 episodes across 710 sessions in 34 scenes, including 1,204 memory-dependent episodes. We also provide MemBridge, a modular baseline that combines semantic scene memory, cross-episode localization, and coarse-to-fine retrieval  to support existing VLN agents. Experiments with  state-of-the-art VLN methods demonstrate the benefits of MemBridge while revealing substantial room for improvement on memory-dependent navigation. The benchmark and baseline code are available at
\url{https://github.com/NJU-R-L-Group-Embodied-Lab/Mem-X}.
\end{abstract}

\section{Introduction}

Embodied agents that operate in real environments over long horizons need to
retain and use observations and experience from past tasks
~\cite{krantz2023ivln,khanna2024goat}.
When an agent continuously executes tasks in the same scene, different tasks
usually share the same locations, and historical memory from preceding
episodes can often assist later tasks~\cite{krantz2023ivln}.
Cross-episode memory is therefore necessary.

Traditional VLN
benchmarks~\cite{anderson2018r2r,ku2020rxr,jain2019r4r,thomason2020cvdn,
qi2020reverie,zhu2021soon} usually evaluate each episode independently, and an
agent's historical information is used only within the current task, making it
difficult to exploit the associations across tasks.

Existing cross-episode VLN
benchmarks~\cite{krantz2023ivln,hong2025gsa} and lifelong navigation
benchmarks already allow agents to retain and use historical information.
However, these benchmarks usually do not make historical memory necessary for
completing later tasks.
Relying only on the spatial relations and actions suggested by the current
instruction, exploration can achieve the same effect.
We therefore consider this task design not tightly coupled with the memory
mechanism, and it cannot truly evaluate an algorithm's cross-episode memory
ability.

\begin{table}[t]
\centering
\caption{Comparison with related benchmarks.}
\label{tab:benchmark_comparison}
\resizebox{\columnwidth}{!}{%
\begin{tabular}{lccc}
\toprule
Benchmark & \shortstack{Cross-Episode\\Memory} &
\shortstack{Memory-\\Dependent} &
\shortstack{Self-\\Localization} \\
\midrule
R2R         & $\times$     & $\times$     & $\times$     \\
GOAT-Bench  & $\times$     & $\times$     & $\times$     \\
IVLN        & $\checkmark$ & $\times$     & $\times$     \\
GSA-R2R     & $\checkmark$ & $\times$     & $\times$     \\
LMEE-Bench  & $\times$     & $\checkmark$ & $\times$     \\
FindingDory & $\checkmark$ & $\checkmark$ & $\times$     \\
\textbf{Ours} & $\boldsymbol{\checkmark}$ & $\boldsymbol{\checkmark}$ & $\boldsymbol{\checkmark}$ \\
\bottomrule
\end{tabular}
}
\end{table}

\begin{figure*}[t]
\centering
\includegraphics[width=\textwidth]{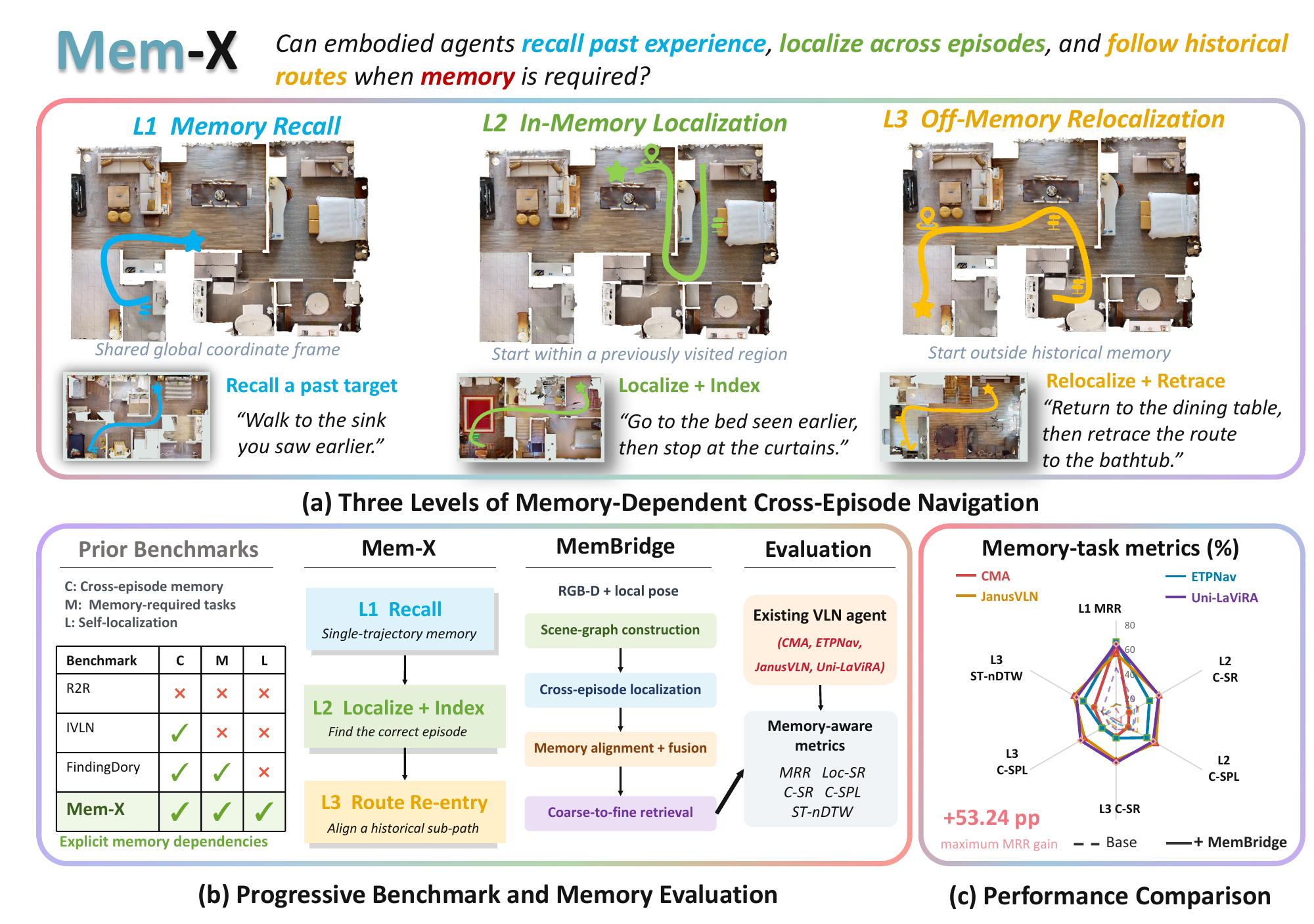}
\caption{Overview of the Mem-X benchmark.
(a) Three memory-dependent levels: In-Trajectory Memory Recall,
In-Memory Localization and Episodic Indexing, and Off-Memory
Relocalization and Route Re-entry.
(b) Mem-X requires historical memory, unlike prior VLN benchmarks;
MemBridge is evaluated with memory-aware metrics.
(c) Memory-task results of four VLN agents with and without MemBridge.}
\label{fig:overview}
\end{figure*}

To address this issue, we construct Mem-X, a cross-episode VLN benchmark for
memory-dependent tasks, as shown in Figure~\ref{fig:overview}.
Its three levels of increasing difficulty are detailed in
Section~\ref{sec:task_design}.

Unlike existing work, we design Mem-X for tightly coupled memory
evaluation.
Through memory recall, in-memory and off-memory localization, episode
indexing, and historical-path reproduction, it forces the agent to use memory
to complete the related tasks.
Table~\ref{tab:benchmark_comparison} compares Mem-X with existing related
benchmarks.

To provide a reference baseline for Mem-X, we introduce MemBridge, a modular approach that combines semantic scene memory, cross-episode localization, and coarse-to-fine retrieval. MemBridge connects historical observations to current navigation tasks and can be integrated with different VLN agents, providing a practical starting point for studying memory-dependent navigation.

The contributions of this paper are summarized as follows:
\begin{itemize}
    \item We introduce Mem-X, a benchmark that makes historical experience an explicit part of VLN task specification through references to previously encountered objects and traversed routes. It comprises 2,071 episodes organized into 710 sessions across 34 scenes, including 1,204 memory-dependent episodes.

    \item We establish an evaluation protocol that controls access to cross-episode coordinates and assesses both memory-related capabilities and task completion, revealing navigation failures that endpoint-based metrics alone overlook.
    
    \item We provide MemBridge, a modular reference baseline that enables existing VLN agents to use historical memory across episodes. Experiments with four representative agents demonstrate its effectiveness while highlighting substantial room for improvement on Mem-X.
\end{itemize}

\section{Related Work}

\subsection{Vision-and-Language Navigation Benchmarks}

VLN benchmarks typically evaluate each episode independently.
R2R~\cite{anderson2018r2r} established the classic setting on the discrete
Matterport3D~\cite{chang2017matterport3d} graph; R4R~\cite{jain2019r4r} and
RxR~\cite{ku2020rxr} extended it to longer and multilingual paths.
CVDN~\cite{thomason2020cvdn}, REVERIE~\cite{qi2020reverie}, and
SOON~\cite{zhu2021soon} study dialog, remote-object, and goal-oriented
variants, while R2R-CE~\cite{krantz2020r2rce} and
LH-VLN~\cite{song2025lhvln} move to continuous and long-horizon settings.
In all of these benchmarks, historical information remains limited to a
single episode.

\begin{figure*}[t]
\centering
\includegraphics[width=\textwidth]{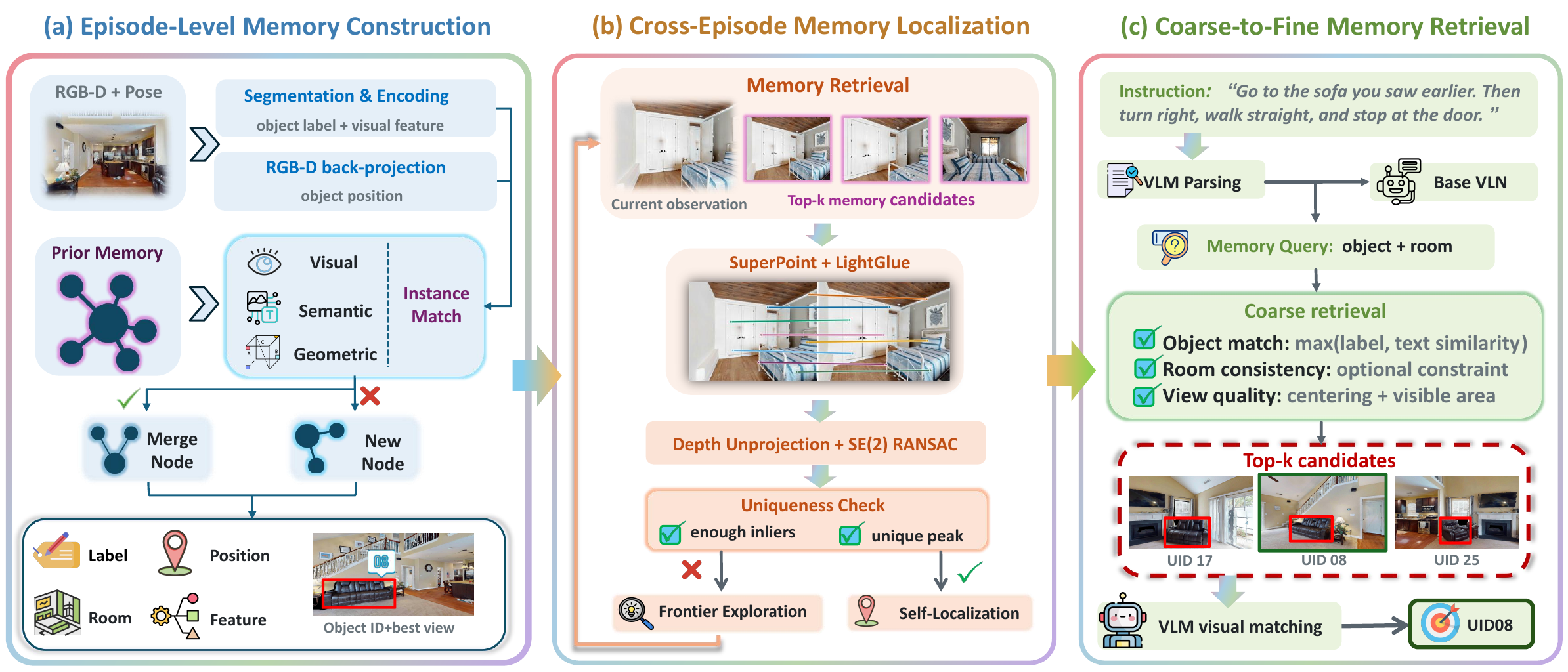}
\caption{Overview of MemBridge, a modular baseline that attaches memory
construction, cross-episode localization, and coarse-to-fine retrieval to an
existing VLN agent.
(a) Episode-Level Memory Construction builds a semantic scene graph from RGB-D observations and local poses.
(b) A transform between episode-local frames aligns new observations with
historical memory, which is then updated.
(c) Coarse-to-fine retrieval identifies the referenced target for planning on
the aligned map.}
\label{fig:method}
\end{figure*}

\subsection{Cross-Episode Memory and Lifelong Embodied Navigation}

In addition to independent single-episode settings, existing studies have also
begun to explore persistent memory across tasks and lifelong embodied
navigation.

IVLN~\cite{krantz2023ivln} organizes independent episodes from existing VLN
datasets into tours that are executed continuously in the same scene, and
allows agents to retain and use historical memory across episodes.
GSA-R2R~\cite{hong2025gsa} evaluates agents that execute a large number of
instructions in the same building and perform scene adaptation.
LENL~\cite{wang2026lenl} is a continual-learning benchmark across scenes and
instruction types.
GOAT-Bench~\cite{khanna2024goat} extends lifelong embodied navigation from the
perspective of multimodal sequential-goal navigation.
These benchmarks usually treat historical memory only as auxiliary information,
and the completion of later tasks does not necessarily depend on historical
memory.

Some recent work tries to emphasize historical memory.
LMEE-Bench~\cite{wang2026lmee} accumulates episodic memory along a single
multi-goal exploration trajectory and then tests recall with question
answering; the memory dependency occurs in QA rather than in later
navigation, and it does not constitute a cross-episode setting.
FindingDory~\cite{yadav2026findingdory} first collects long-horizon
interactions, then retrieves a target frame from the historical video and
navigates back, which evaluates in-trajectory memory recall quite
thoroughly.
Its history records include pose, and the agent is also equipped with GPS,
so it does not require cross-episode self-localization and memory alignment
without a shared global pose.
Mem-X instead organizes multiple episodes into sessions with explicit
dependencies and adds in-memory / off-memory localization.

\subsection{Memory-Augmented VLN Agents}

Memory in existing VLN methods is typically implicit temporal memory or
explicit spatial memory.
Implicit methods encode history inside the model, including recurrent hidden
states ~\cite{krantz2020r2rce}, Transformer-based history encoding and
caching
~\cite{chen2021hamt,lin2021mtvm,zeng2025janusvln}, state-space recurrent
memory ~\cite{zhang2025cosmo,yao2025navmorph}, and
multi-timescale memory
~\cite{wei2026streamvln,liu2026memvln,zhang2025uninavid}.

Explicit methods store memory in external structures, including topological
and scene graphs
~\cite{an2024etpnav,chen2022duet,zhou2025fsrvln}, metric or semantic maps
~\cite{wang2023gridmm,zhang2025mapnav,wang2024vln3dff}, and waypoint-indexed
episodic memory ~\cite{ding2026unilavira,xu2025memoir}.
These mechanisms are mainly used within a single episode, or accumulate maps
with pose, and are not designed for memory-dependent tasks across episodes
without a shared global pose.

\section{Benchmark}

\subsection{Task Design}
\label{sec:task_design}

Mem-X groups navigation episodes into sessions. Agents accumulate experience during initial tasks and use it in subsequent tasks, whose instructions refer to previously encountered objects or traversed routes. 

We design three levels that progressively extend historical-object recall to cross-episode localization and historical-route reproduction. All levels provide drift-free poses derived from simulator ground truth. Level 1 uses a shared coordinate frame across episodes to isolate historical-object recall from cross-episode localization. Levels 2 and 3 instead express poses in episode-local frames obtained by resetting the simulator pose. Transformations between episode frames are withheld, requiring agents to establish spatial correspondence with historical observations.

\textbf{Level 1: In-Trajectory Memory Recall.}
Each session consists of a base episode, Episode 0, followed by 1–4 memory-dependent tasks. These tasks start at the endpoint of the trajectory actually executed in Episode 0 or at a point sampled from its 75\%–90\% interval. The agent must identify and return to an object referenced through the preceding experience. By placing the start on the historical trajectory and providing a shared coordinate frame, this level focuses on memory construction and target recall while reducing the need for unexplored-region navigation.

\textbf{Level 2: In-Memory Localization and Episodic Indexing.}
Each session contains three episodes with separate local coordinate frames. Episode 0 provides the initial experience. Episode 1 starts 1.0–1.5 m from the start of the Episode 0 reference trajectory and requires the agent to revisit a target from Episode 0 before following a new navigation instruction. Episode 2 starts at the same position as Episode 1 and refers to a target encountered in Episode 1 but not Episode 0. The agent must relate its current observations to the visited region and retrieve the target associated with the relevant episode.

\textbf{Level 3: Off-Memory Relocalization and Route Re-entry.}
Each session contains two base episodes whose reference trajectories share a continuous segment, followed by a memory-dependent episode. All three episodes use separate local coordinate frames. The final episode starts outside historical observation coverage, requiring the agent to explore and establish correspondence with a previously visited region. It must then identify and reproduce a specified sub-path of the shared historical segment before completing the remaining instruction. Specifying a sub-path requires selecting the relevant portion of the historical route, beyond simply returning to a familiar region. This level evaluates off-memory relocalization and ordered historical-route reproduction.

Memory-related instructions refer to past experience without directly describing how to reach the target. Agents must visit the specified targets or follow the required historical routes, but may use any memory representation or navigation method.

\subsection{Manual Trajectory Collection}

We collect reference trajectories and their corresponding navigation instructions in HM3DSem scenes ~\cite{yadav2023hm3dsem} using the Habitat simulator ~\cite{savva2019habitat}. Collection is organized at the session level so that later tasks can refer to objects and routes encountered in earlier episodes. Collectors first familiarize themselves with the scene, then design the instructions and reference paths according to the level-specific requirements in
Section~\ref{sec:task_design}. They manually control the agent to execute each planned route, producing paired instructions and trajectories.

The agent's forward step size is $0.25\,\mathrm{m}$, and each turn is
$15^{\circ}$.
The agent and the RGB-D sensor are both $1.2\,\mathrm{m}$ high.
The image resolution is $960\times720$, with a $90^{\circ}$ horizontal field
of view.
During collection, we record the complete action sequence, 3D position,
orientation, task start and end points, and the reference path.
These recordings are used for data generation and cleaning in
Section~\ref{sec:data_gen}.

\subsection{Data Generation and Cleaning}
\label{sec:data_gen}

\textbf{Level 1:} For each collected trajectory, we manually annotate visible
and distinctive anchor objects along the route and use them as memory recall
targets for subsequent episodes.

\textbf{Level 2:} To prevent the agent from directly discovering a target in
the initial observation and bypassing self-localization and memory fusion, the
respective recall targets of Episode 1 and Episode 2 are not visible at their
starting points, and their shortest geodesic distances from the corresponding
starting point are no less than 3.25~m.

\textbf{Level 3:} We keep Episode~0--1 pairs whose overlapping sub-paths have
bidirectional coverage of at least $95\%$ and bidirectional P95 and ordered
discrete Fr\'{e}chet distances of at most $1.2\,\mathrm{m}$; the
$0.6\,\mathrm{m}$ neighborhoods of this overlap define the shared route.
Episode~2 must follow that route: $95\%$ of its points lie within
$0.8\,\mathrm{m}$ of either memory trajectory, the P95 distance is at most
$0.8\,\mathrm{m}$, and the maximum deviation is at most $1.2\,\mathrm{m}$.
It starts at least $1.0\,\mathrm{m}$ geodesic from both.
RGB-D replay of the first two episodes further removes starts observed in any
historical frame.

Finally, we manually review all samples and remove those that do not satisfy the
task constraints or contain annotation errors to ensure the accuracy and
reliability of the dataset.

\subsection{Data Statistics}

As shown in Table~\ref{tab:dataset_statistics}, we construct a total of 710
sessions, among which Level 1, Level 2, and Level 3 contain 315, 238, and 157
sessions, respectively. The dataset contains 2,071 VLN trajectories in total,
of which 1,204 are memory-dependent episodes, covering 34 scenes and 110 classes
of memory recall targets.

\begin{table}[t]
\centering
\caption{Statistics of Mem-X by level.}
\label{tab:dataset_statistics}
\resizebox{\columnwidth}{!}{%
\begin{tabular}{lcccc}
\toprule
Level & Episodes & Memory-Dependent Episodes & Avg.\ path length & Avg.\ steps \\
\midrule
L1  & 886   & 571  & $9.86 \pm 4.99$ m  & 60  \\
L2  & 714   & 476  & $16.52 \pm 6.00$ m & 99  \\
L3  & 471   & 157  & $20.50 \pm 5.66$ m & 106 \\
All & 2,071 & 1,204 & $14.58 \pm 7.01$ m & 84  \\
\bottomrule
\end{tabular}
}
\end{table}

\subsection{Evaluation Protocol}
\label{sec:eval_protocol}

We establish an evaluation protocol to systematically assess Mem-X.
The protocol controls access to cross-episode coordinates
(Section~\ref{sec:task_design}): Level~1 shares a global frame, while
Levels~2 and~3 provide only episode-local poses and withhold inter-episode
transforms, so later tasks cannot be solved from a shared pose.
Following common VLN settings~\cite{anderson2018r2r,krantz2020r2rce}, we
report Success Rate (SR) and Success weighted by Path Length (SPL) to measure
task completion.
Because these endpoint-based metrics cannot tell whether the agent used
historical memory, we further propose metrics for memory-related
capabilities, covering target recall, cross-episode localization, constrained
completion, and historical-route reproduction.

Because we cannot accurately measure whether the agent truly found the target
object, we use Memory Recall Rate (MRR) as a coarse measure of memory-target
recall accuracy.
A memory task is recalled if any actual state is within $1\,\mathrm{m}$
geodesic of the query target $g_i$:
\begin{equation}
\mathrm{MRR}
=
\frac{1}{N}
\sum_{i=1}^{N}
\mathbb{I}\!\left(
\min_{t}
d_{\mathrm{geo}}(s_t,g_i)
\le 1\,\mathrm{m}
\right).
\end{equation}

For cross-episode localization at Level~2 and Level~3, we use Localization
Success Rate (Loc-SR).
The evaluated set includes Level~2 memory tasks and, at Level~3, both
Episode~1 and the memory episode.
Although Episode~1 is a base task, it still requires localization to align
the current observation with historical memory.
Let the pose estimate of the $i$-th evaluated task be
$\mathbf{T}_i=(\mathbf{p}_i,\theta_i)$ with Habitat ground-plane
$\mathbf{p}=(x,z)$ in meters.
The translation and heading errors are
\begin{equation}
\begin{aligned}
E_{\mathrm{trans}}^{(i)}
&=
\sqrt{(x_i-x_i^{\mathrm{gt}})^2+(z_i-z_i^{\mathrm{gt}})^2},
\\
E_{\mathrm{rot}}^{(i)}
&=
\bigl|
\operatorname{wrap}(\theta_i-\theta_i^{\mathrm{gt}})
\bigr|,
\end{aligned}
\end{equation}
where
$\operatorname{wrap}(\Delta\theta)
=\bigl((\Delta\theta+\pi)\bmod 2\pi\bigr)-\pi$
and $E_{\mathrm{rot}}\in[0,\pi]$, i.e., $[0^\circ,180^\circ]$.
Tasks without a pose estimate are treated as localization failures.
With threshold $0.25\,\mathrm{m}$ / $10^\circ$,
\begin{equation}
\mathrm{Loc\text{-}SR}
=
\frac{1}{N}
\sum_{i=1}^{N}
\mathbb{I}\!\left(
E_{\mathrm{trans}}^{(i)}\le 0.25\,\mathrm{m}
\wedge
E_{\mathrm{rot}}^{(i)}\le 10^\circ
\right).
\end{equation}
Mean errors are computed only on tasks that produce a pose estimate,
$\mathcal{L}$:
$\bar{E}_{\mathrm{trans}}
=\frac{1}{|\mathcal{L}|}\sum_{i\in\mathcal{L}}E_{\mathrm{trans}}^{(i)}$
and
$\bar{E}_{\mathrm{rot}}
=\frac{1}{|\mathcal{L}|}\sum_{i\in\mathcal{L}}E_{\mathrm{rot}}^{(i)}$.

Memory tasks at Level~2 and Level~3 require the agent to first recall the
specified memory and then reach the final goal.
We therefore define Constrained Success Rate (C-SR) and Constrained SPL
(C-SPL):
\begin{equation}
\mathrm{C\text{-}SR}
=\frac{1}{N}\sum_{i=1}^{N}C_i,
\qquad
\mathrm{C\text{-}SPL}
=\frac{1}{N}\sum_{i=1}^{N}
C_i\frac{L_i^C}{\max(L_i^C,P_i)}.
\end{equation}
Here $C_i=1$ if the agent reaches the specified memory anchors in order and
finally completes the task, and $0$ otherwise; $P_i$ is the actual path length,
and $L_i^C$ is the shortest path length that satisfies the constraint.

Level~3 further requires the agent to reproduce a historical overlapping
route, not only to reach the referenced region.
Following the standard nDTW definition~\cite{ilharco2019ndtw},
\begin{equation}
\begin{aligned}
\mathrm{nDTW}(\tau,\tau^{*})
&=
\exp\!\left(
-\frac{\mathrm{DTW}(\tau,\tau^{*})}{|\tau^{*}|\,d_{\mathrm{th}}}
\right),
\\
\mathrm{ST\text{-}nDTW}
&=
\frac{1}{N}
\sum_{i=1}^{N}
\max_{k}
\mathrm{nDTW}\!\left(\tau_i^{\mathrm{rev}},\tau_{i,k}^{\mathrm{ov}}\right),
\end{aligned}
\end{equation}
where $\tau$ is the query path, $\tau^{*}$ is the reference path,
$\mathrm{DTW}$ is the dynamic time warping distance, and $d_{\mathrm{th}}$ is
the success-distance threshold.
ST-nDTW is the mean, over Level~3 memory tasks, of the maximum nDTW between
the revisit segment $\tau_i^{\mathrm{rev}}$ and the GT overlapping
sub-trajectories $\tau_{i,k}^{\mathrm{ov}}$ of the preceding base episodes.
It measures whether the agent completes the revisit along at least one
historically overlapping route.

\section{MemBridge}

\subsection{Episode-Level Memory Construction}

As shown in Figure~\ref{fig:method}(a), MemBridge incrementally constructs a
semantic scene
graph~\cite{werby2024hovsg,jiang2025dualmap,gu2023conceptgraphs,fang2026inheritsg}
within a single episode from the current RGB-D observation and the local pose.

Keyframes are selected with DINOv3~\cite{simeoni2025dinov3} features when the
similarity to the previous keyframe falls below a threshold.
For each keyframe, SAM3~\cite{carion2026sam3} detects and segments objects,
DINOv3 extracts region features, and three-dimensional positions are computed
from depth and camera poses.
Following INHerit-SG~\cite{fang2026inheritsg}, a new observation is associated
with existing nodes by visual-feature similarity, semantic consistency, and
geometric distance; a matched instance is updated, otherwise a new node is
created.
Object nodes also store an episode index for later retrieval.

\subsection{Cross-Episode Memory Localization}

As shown in Figure~\ref{fig:method}(b), MemBridge first performs
self-localization across episodes.
The current observation is encoded with a DINOv3~\cite{simeoni2025dinov3}
global feature, compared by cosine similarity to historical keyframes, and
the top-$k$ frames are selected.

SuperPoint~\cite{detone2018superpoint} and
LightGlue~\cite{lindenberger2023lightglue} establish correspondences between
the current RGB image and these $k$ frames.
Each match is back-projected from depth and projected onto the ground plane.
RANSAC~\cite{fischler1981ransac} then estimates a planar transform
$g\in\mathrm{SE}(2)$ on each candidate.

Because similar appearance at different places can make $g$ unreliable, we
accept it as the unique valid transform only when the best candidate has at
least 80 inliers and, if the second-best has at least 20 inliers, the best is
still no less than $1.5\times$ the second-best.
A valid $g$ warps the current observations into the historical scene-graph
frame and updates the graph.
If no unique $g$ is found, a frontier on the agent's occupancy map whose
heading is most similar to a historical frame is selected, and
self-localization is repeated every $1\,\mathrm{m}$.

\subsection{Coarse-to-Fine Memory Retrieval}

As shown in Figure~\ref{fig:method}(c), MemBridge uses coarse-to-fine
two-stage retrieval over the scene graph.
A VLM parses the instruction into segments and routes each segment to memory
retrieval or the base VLN model.
For a memory-recall segment, the coarse score of each object node is the
maximum of label matching and text similarity to the target category.
Nodes are ranked by this score, room consistency, and historical observation
quality, and the Top-$k$ candidates are kept.

In the fine stage, the original instruction, these $k$ candidates, and their
historical keyframes are input to the VLM, which selects the target UID from
appearance and scene context.
If no suitable candidate exists, recall is rejected and the system falls back
to the base VLN model.
After a target is selected, its stored position is used for planning on the
aligned occupancy map.

\begin{figure*}[t]
\centering
\includegraphics[width=\textwidth]{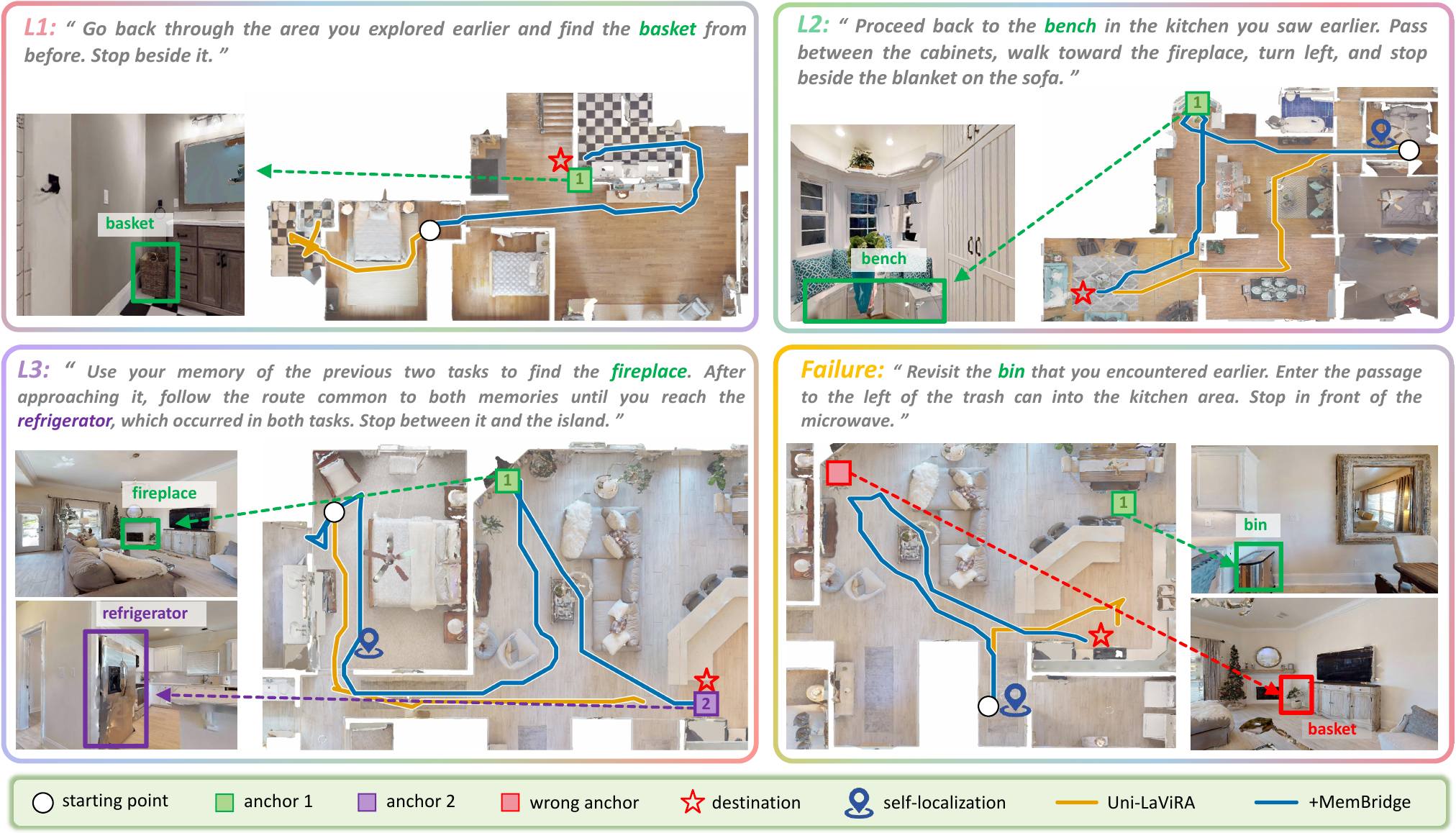}
\caption{Qualitative cases of Uni-LaViRA with and without MemBridge:
one case for each of the three levels, plus a MemBridge failure case.
Yellow denotes the native Uni-LaViRA trajectory, and blue denotes the
Uni-LaViRA+MemBridge trajectory.}
\label{fig:qualitative}
\end{figure*}

\section{Experiments}

\subsection{Implementation Details}
\label{sec:impl}

We conduct experiments on eight NVIDIA RTX 4090 GPUs.
Instruction parsing and Top-$k$ fine retrieval use Qwen-3.5-27B, with VLM
fine-retrieval Top-$k$ of 3 and historical-keyframe retrieval Top-$k$ of 5.
The keyframe-selection threshold for semantic-graph construction is 0.85, the
visual-similarity threshold is 0.80, and the spatial-distance threshold is
$1.5\,\mathrm{m}$.
The text encoder is all-MiniLM-L6-v2~\cite{wang2020minilm}.
All methods at Level~2 and Level~3 use simulator-obtained local poses.

We evaluate four agents with different memory mechanisms.
CMA~\cite{krantz2020r2rce} uses a recurrent GRU, ETPNav~\cite{an2024etpnav}
uses a topological graph, JanusVLN~\cite{zeng2025janusvln} uses a Transformer
history cache, and Uni-LaViRA~\cite{ding2026unilavira} uses waypoint-indexed
episodic memory.
CMA's GRU is cleared each episode, because carrying the hidden state causes
an immediate STOP.
JanusVLN does not depend on pose, so its cache is retained at all three
levels.
ETPNav and Uni-LaViRA use position-dependent inputs: native memory is
retained at Level~1, which uses a global frame, and is reset at Levels~2
and~3, which use episode-local frames, to avoid misaligned historical poses.

\subsection{Results Analysis}

As shown in Table~\ref{tab:main_results}, the baseline results reveal
different limitations across agents.
At Level~1, Uni-LaViRA achieves the highest baseline MRR of $41.68\%$,
compared with $12.43\%$ for JanusVLN and $12.78\%$ for ETPNav, even though
JanusVLN is stronger on the base navigation tasks.

\begin{table*}[t]
\centering
\caption{Experimental results of four methods on Mem-X.
Each method and its corresponding +MemBridge variant share the same Basic results and are compared only on memory tasks.
+MemBridge denotes the percentage-point improvement on memory tasks
after inserting the MemBridge module.}
\label{tab:main_results}
\setlength{\tabcolsep}{2.6pt}
\renewcommand{\arraystretch}{1.12}
\newcommand{\gplus}[1]{\cellcolor[gray]{0.92}#1}
\resizebox{\textwidth}{!}{%
\begin{tabular}{@{}llcccccccccccccc@{}}
\toprule
Method &
& \multicolumn{3}{c}{Level 1}
& \multicolumn{5}{c}{Level 2}
& \multicolumn{6}{c}{Level 3} \\
\cmidrule(lr){3-5}\cmidrule(lr){6-10}\cmidrule(lr){11-16}
&
& SR $\uparrow$ & SPL $\uparrow$ & MRR $\uparrow$
& SR $\uparrow$ & SPL $\uparrow$ & C-SR $\uparrow$ & C-SPL $\uparrow$ & MRR $\uparrow$
& SR $\uparrow$ & SPL $\uparrow$ & C-SR $\uparrow$ & C-SPL $\uparrow$
& ST-nDTW $\uparrow$ & MRR $\uparrow$ \\
\midrule
\multirow{3}{*}{CMA} & Basic
& 16.19\% & 15.51\% & --
& 29.41\% & 28.14\% & -- & -- & --
& 7.96\% & 7.58\% & -- & -- & -- & -- \\
& Memory
& 18.91\% & 17.12\% & 8.93\%
& 9.87\% & 9.44\% & 4.83\% & 4.66\% & 8.19\%
& 3.82\% & 3.81\% & 0.00\% & 0.00\% & 9.14\% & 5.41\% \\
& \textit{+MemBridge}
& \gplus{$+48.52\%$} & \gplus{$+43.46\%$} & \gplus{$+53.24\%$}
& \gplus{$+6.73\%$} & \gplus{$+5.84\%$} & \gplus{$+6.51\%$} & \gplus{$+6.36\%$} & \gplus{$+14.08\%$}
& \gplus{$+7.01\%$} & \gplus{$+5.58\%$} & \gplus{$+5.73\%$} & \gplus{$+5.49\%$} & \gplus{$+4.73\%$} & \gplus{$+9.56\%$} \\
\hdashline
\multirow{3}{*}{ETPNav} & Basic
& 15.56\% & 13.18\% & --
& 57.98\% & 51.82\% & -- & -- & --
& 28.98\% & 25.58\% & -- & -- & -- & -- \\
& Memory
& 19.96\% & 15.89\% & 12.78\%
& 25.42\% & 21.49\% & 19.12\% & 16.01\% & 28.36\%
& 10.19\% & 10.10\% & 0.00\% & 0.00\% & 13.14\% & 8.92\% \\
& \textit{+MemBridge}
& \gplus{$+48.87\%$} & \gplus{$+46.98\%$} & \gplus{$+50.09\%$}
& \gplus{$+9.87\%$} & \gplus{$+9.37\%$} & \gplus{$+10.08\%$} & \gplus{$+10.66\%$} & \gplus{$+7.35\%$}
& \gplus{$+16.56\%$} & \gplus{$+9.93\%$} & \gplus{$+10.19\%$} & \gplus{$+7.97\%$} & \gplus{$+5.04\%$} & \gplus{$+23.57\%$} \\
\hdashline
\multirow{3}{*}{JanusVLN} & Basic
& 49.21\% & 48.08\% & --
& 72.27\% & 70.33\% & -- & -- & --
& 63.69\% & 55.34\% & -- & -- & -- & -- \\
& Memory
& 43.08\% & 37.11\% & 12.43\%
& 34.87\% & 30.05\% & 14.08\% & 11.95\% & 18.28\%
& 28.03\% & 24.64\% & 5.73\% & 5.67\% & 13.80\% & 19.43\% \\
& \textit{+MemBridge}
& \gplus{$+23.65\%$} & \gplus{$+24.25\%$} & \gplus{$+41.16\%$}
& \gplus{$+11.13\%$} & \gplus{$+8.90\%$} & \gplus{$+22.48\%$} & \gplus{$+19.10\%$} & \gplus{$+32.14\%$}
& \gplus{$+10.82\%$} & \gplus{$+4.36\%$} & \gplus{$+22.93\%$} & \gplus{$+20.36\%$} & \gplus{$+19.34\%$} & \gplus{$+27.07\%$} \\
\hdashline
\multirow{3}{*}{Uni-LaViRA} & Basic
& 36.51\% & 29.86\% & --
& 62.61\% & 47.89\% & -- & -- & --
& 36.62\% & 28.96\% & -- & -- & -- & -- \\
& Memory
& 63.57\% & 48.38\% & 41.68\%
& 33.61\% & 28.05\% & 14.50\% & 12.12\% & 19.54\%
& 27.39\% & 21.23\% & 7.64\% & 6.80\% & 13.15\% & 20.38\% \\
& \textit{+MemBridge}
& \gplus{$+10.51\%$} & \gplus{$+19.81\%$} & \gplus{$+19.44\%$}
& \gplus{$+12.40\%$} & \gplus{$+8.03\%$} & \gplus{$+27.73\%$} & \gplus{$+20.63\%$} & \gplus{$+28.78\%$}
& \gplus{$+11.46\%$} & \gplus{$+8.13\%$} & \gplus{$+19.11\%$} & \gplus{$+17.73\%$} & \gplus{$+15.75\%$} & \gplus{$+22.30\%$} \\
\bottomrule
\end{tabular}%
}
\end{table*}

\begin{table}[t]
\centering
\caption{Localization results of MemBridge-augmented methods.}
\label{tab:loc_sr}
\setlength{\tabcolsep}{3.2pt}
\renewcommand{\arraystretch}{1.12}
\resizebox{\columnwidth}{!}{%
\begin{tabular}{@{}lcccccc@{}}
\toprule
& \multicolumn{3}{c}{Level 2}
& \multicolumn{3}{c}{Level 3} \\
\cmidrule(lr){2-4}\cmidrule(lr){5-7}
Method
& Loc-SR $\uparrow$ & $E_{\mathrm{trans}}$ $\downarrow$ & $E_{\mathrm{rot}}$ $\downarrow$
& Loc-SR $\uparrow$ & $E_{\mathrm{trans}}$ $\downarrow$ & $E_{\mathrm{rot}}$ $\downarrow$ \\
\midrule
CMA\textit{+MemBridge}
& 97.5\% & $0.037\,\mathrm{m}$ & $0.78^{\circ}$
& 62.74\% & $0.181\,\mathrm{m}$ & $2.26^{\circ}$ \\
ETPNav\textit{+MemBridge}
& 98.3\% & $0.026\,\mathrm{m}$ & $0.34^{\circ}$
& 74.20\% & $0.122\,\mathrm{m}$ & $0.93^{\circ}$ \\
JanusVLN\textit{+MemBridge}
& 99.2\% & $0.022\,\mathrm{m}$ & $0.56^{\circ}$
& 92.36\% & $0.174\,\mathrm{m}$ & $2.02^{\circ}$ \\
Uni-LaViRA\textit{+MemBridge}
& 98.1\% & $0.021\,\mathrm{m}$ & $0.33^{\circ}$
& 70.70\% & $0.310\,\mathrm{m}$ & $2.43^{\circ}$ \\
\bottomrule
\end{tabular}%
}
\end{table}

\begin{table*}[t]
\centering
\caption{Ablation of MemBridge components on Uni-LaViRA memory tasks.}
\label{tab:ablation}
\setlength{\tabcolsep}{2.6pt}
\renewcommand{\arraystretch}{1.12}
\resizebox{\textwidth}{!}{%
\begin{tabular}{@{}lcccccccccccccc@{}}
\toprule
Method
& \multicolumn{3}{c}{Level 1}
& \multicolumn{5}{c}{Level 2}
& \multicolumn{6}{c}{Level 3} \\
\cmidrule(lr){2-4}\cmidrule(lr){5-9}\cmidrule(lr){10-15}
& SR $\uparrow$ & SPL $\uparrow$ & MRR $\uparrow$
& SR $\uparrow$ & SPL $\uparrow$ & C-SR $\uparrow$ & C-SPL $\uparrow$ & MRR $\uparrow$
& SR $\uparrow$ & SPL $\uparrow$ & C-SR $\uparrow$ & C-SPL $\uparrow$
& ST-nDTW $\uparrow$ & MRR $\uparrow$ \\
\midrule
\textbf{Uni-LaViRA\textit{+MemBridge}}
& \textbf{74.08\%} & \textbf{68.19\%} & \textbf{61.12\%}
& \textbf{46.01\%} & \textbf{36.08\%} & \textbf{42.23\%} & \textbf{32.75\%} & \textbf{48.32\%}
& \textbf{38.85\%} & \textbf{29.36\%} & \textbf{26.75\%} & \textbf{24.53\%}
& \textbf{28.90\%} & \textbf{42.68\%} \\
w/o VLM Top-$k$ fine retrieval
& 63.40\% & 58.16\% & 52.71\%
& 38.87\% & 30.50\% & 35.71\% & 27.66\% & 41.81\%
& 28.66\% & 21.88\% & 19.75\% & 17.82\%
& 21.30\% & 31.21\% \\
w/o cross-episode localization
& 61.82\% & 45.11\% & 39.93\%
& 34.03\% & 29.66\% & 15.69\% & 13.41\% & 19.75\%
& 26.11\% & 20.28\% & 7.52\% & 5.73\%
& 11.90\% & 20.70\% \\
w/o frontier exploration
& 73.38\% & 66.74\% & 59.54\%
& 44.11\% & 36.08\% & 40.76\% & 31.83\% & 45.80\%
& 28.66\% & 21.96\% & 11.46\% & 10.32\%
& 15.80\% & 24.52\% \\
\bottomrule
\end{tabular}%
}
\end{table*}

At Levels~2 and~3, following the memory-handling protocol in
Section~\ref{sec:impl}, JanusVLN retains its history cache but still reaches
only $5.73\%$ C-SR at Level~3 against an SR of $28.03\%$.

Similar SR--C-SR gaps appear across agents.
Reaching the destination does not guarantee visiting the required historical
anchors in order.

Adding MemBridge improves performance across the evaluated agents.
At Level~3, C-SR increases from $5.73\%$ to $28.66\%$ for JanusVLN and from
$7.64\%$ to $26.75\%$ for Uni-LaViRA; ST-nDTW increases from $13.80\%$ to
$33.14\%$ and from $13.15\%$ to $28.90\%$.
Table~\ref{tab:loc_sr} reports localization on 476 Level~2 memory tasks
and 314 Level~3 tasks.
Although Episode~1 at Level~3 is a base task, it still requires
localization to align the current observation with historical memory.
At Level~2, JanusVLN and Uni-LaViRA with MemBridge reach Loc-SR of $99.2\%$
and $98.1\%$, while C-SR remains $36.56\%$ and $42.23\%$, so spatial
alignment alone is insufficient.

\subsection{Ablation Study}
\label{sec:ablation}

We analyze the components of MemBridge to examine how different parts of the
reference baseline contribute to performance on Mem-X.
Table~\ref{tab:ablation} reports the effects of removing VLM Top-$k$ fine
retrieval, cross-episode localization, and frontier exploration.

Removing VLM Top-$k$ fine retrieval reduces Level~1 SR from $74.08\%$ to
$63.40\%$ and MRR from $61.12\%$ to $52.71\%$.
Level~2 and Level~3 also decline, but the largest drop is on Level~1, where
the shared frame already isolates target recall from localization.
Coarse scene-graph ranking is therefore insufficient when several candidates
share a category.

Removing cross-episode localization yields metrics close to Uni-LaViRA's
native Memory setting, effectively reducing the agent to in-episode memory.

Frontier exploration is used only when the first observation does not yield
a unique lock.
Removing it leaves Level~1 and Level~2 largely unchanged, because Level~1
does not require cross-episode localization and Level~2 starts in a
previously observed region.
On Level~3, SR falls from $38.85\%$ to $28.66\%$ and C-SR from $26.75\%$ to
$11.46\%$, approaching the native Memory row.
Off-memory relocalization therefore depends on exploring into historical
coverage, not on a start-view match.

\subsection{Qualitative Analysis}
\label{sec:qual}

We select Uni-LaViRA, which has stronger overall performance, and analyze
the four cases shown in Figure~\ref{fig:qualitative}.

The Level~1 case requires returning to a previously seen basket.
Uni-LaViRA does not recall this object, and its trajectory misses the
basket; after MemBridge is added, the agent stops beside it.

The Level~2 case uses local coordinates and requires first returning to a
previously seen bench, then stopping beside a blanket.
Uni-LaViRA's trajectory cuts directly toward the goal, with SR$=1$ and
C-SR$=0$: the memory task is reduced to a final ordinary-navigation
segment.
MemBridge completes self-localization at the start and then visits the two
anchors in order.

The Level~3 case starts from a region not covered by history.
The instruction requires first finding a previously seen fireplace, then
following the shared route of two memory segments to a refrigerator.
Uni-LaViRA can approach the goal (SR$=1$) but does not visit the
anchors in order; ST-nDTW is near $0$, and the shared route is barely
reproduced.
MemBridge first establishes spatial correspondence with the historical
region during exploration, then visits the two anchors in sequence.

The failure case requires first revisiting a trash bin, then entering the
kitchen and stopping in front of a microwave.
MemBridge has already locked the pose, but retrieves the instruction's bin
as a basket in the same scene, yielding C-SR$=0$.

\FloatBarrier
\section{Conclusion}

We propose Mem-X, a cross-episode memory-dependent VLN benchmark with explicit
dependencies between tasks.
We also introduce an evaluation protocol that measures memory-related
capabilities beyond endpoint success.
MemBridge is a modular baseline that combines semantic scene memory,
cross-episode localization, and coarse-to-fine retrieval and can be
integrated with existing VLN agents.
Experiments with four VLN agents show that MemBridge improves memory-related
metrics, but reaching the goal or aligning poses does not mean the agent used
historical memory.
Even with MemBridge, Level~3 constrained success remains below $30\%$, leaving
substantial room for improvement.
The current dataset covers only single-floor environments, and the odometry
is ground-truth from the simulator, which still differs from real settings.
Future work can extend the benchmark to multi-floor episodes and more
realistic task designs.

\bibliographystyle{IEEEtran}
\bibliography{references}

\end{document}